\pdfoutput=1

\documentclass[11pt]{article}

\usepackage[]{ACL2023}

\usepackage{times}
\usepackage{latexsym}

\usepackage[T1]{fontenc}

\usepackage[utf8]{inputenc}

\usepackage{microtype}

\usepackage{inconsolata}
\usepackage{amsfonts}
\usepackage{amsmath}
\usepackage{graphicx}
\usepackage{multirow}
\usepackage{makecell}
\usepackage{caption}
\usepackage{subfigure}
\usepackage{enumitem}
\usepackage{float}
\usepackage{makecell}
\usepackage{booktabs}
\usepackage{tabu}
\usepackage{tabularx}
\usepackage{pifont}
\usepackage{xcolor}

\definecolor{radarblue}{RGB}{126,166,224}
\definecolor{radarpink}{RGB}{234,107,102}

\title{Context-aware Adversarial Attack on Named Entity Recognition}

\author{
    Shuguang Chen \\ University of Houston \\ \texttt{schen52@uh.edu} \\
    \And
    Leonardo Neves \\ Snap Inc. \\ \texttt{lneves@snap.com} \\
    \And
    Thamar Solorio \\ University of Houston \\ 
    Mohamed bin Zayed University\\ of Artificial Intelligence\\
    \texttt{tsolorio@uh.edu} \\
}

\begin{document}
\maketitle

\begin{abstract}
In recent years, large pre-trained language models (PLMs) have achieved remarkable performance on many natural language processing benchmarks. Despite their success, prior studies have shown that PLMs are vulnerable to attacks from adversarial examples. In this work, we focus on the named entity recognition task and study context-aware adversarial attack methods to examine the model's robustness. Specifically, we propose perturbing the most informative words for recognizing entities to create adversarial examples and investigate different candidate replacement methods to generate natural and plausible adversarial examples. Experiments and analyses show that our methods are more effective in deceiving the model into making wrong predictions than strong baselines.
\end{abstract}

\section{Introduction}

Existing methods for adversarial attacks mainly focus on text classification \citep{DBLP:conf/ijcai/0002LSBLS18, garg-ramakrishnan-2020-bae}, machine translation \citep{DBLP:conf/iclr/BelinkovB18, cheng-etal-2019-robust}, reading comprehension \citep{jia-liang-2017-adversarial, wallace-etal-2019-universal}, etc. A slight perturbation to the input can deceive the model into making wrong predictions or leaking important information. Such adversarial attacks are widely used to identify potential vulnerabilities and audit the model robustness. However, in the context of named entity recognition (NER), these adversarial attack methods are inadequate since they are not customized for the labeling schemes in NER \citep{lin-etal-2021-rockner}. This is especially problematic as the generated adversarial examples can be mislabeled.

\begin{figure}[t]
\centering
\includegraphics[width=0.9\linewidth]{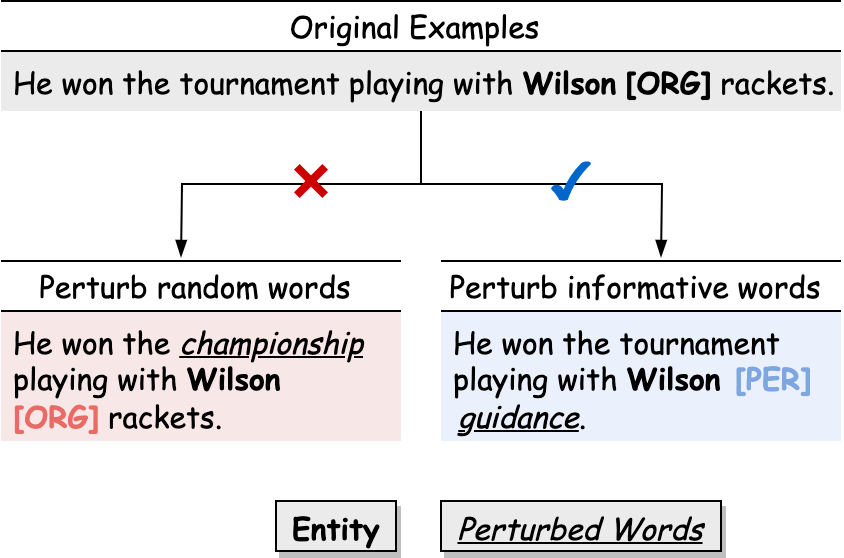}
\caption{Comparison between adversarial attack with and without perturbing informative words.}
\label{fig: illustration_examples}
\end{figure}

Prior studies have proposed various context-aware attacks (i.e., perturb non-entity words) and entity attack (i.e., perturb only entity words) methods to address this issue. Despite their success, most existing methods randomly select words to perturb without taking the linguistic structure into consideration, limiting their effectiveness to consistently generate natural and coherent adversarial examples. Some words in a sentence are more informative than others in guiding the model to recognize named entities. For instance, in Figure \ref{fig: illustration_examples}, the word ``rackets" can provide more information than the word ``tournament" to infer the entity type of ``Wilson". Perturbing such words can be effective in leading to more incorrect model predictions.

\begin{figure*}[ht]
\centering
\includegraphics[width=1.0\linewidth]{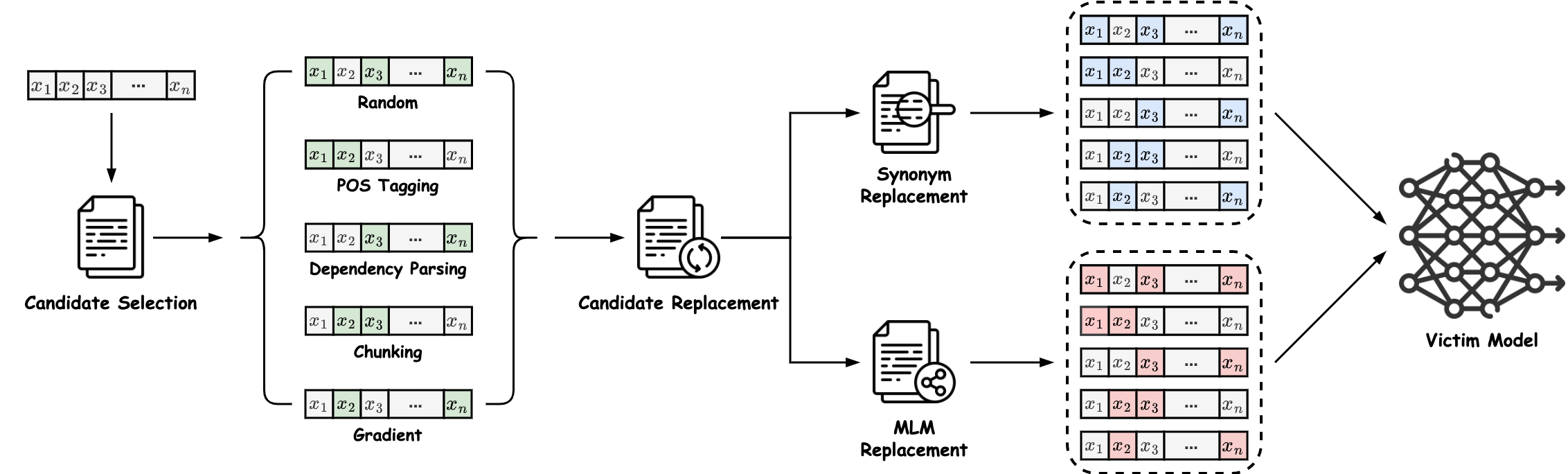}
\caption{The pipeline of the proposed context-aware adversarial attack, including candidate selection to determine which words to perturb and candidate replacement for replacing candidate words.}
\label{fig: pipeline}
\end{figure*}

In this work, we explore the correlation between model vulnerability and informative words. We aim to conduct adversarial attacks by perturbing the informative words to expose the potential vulnerabilities of NER systems. To this end, we investigate different candidate selection methods to determine which words should be perturbed, including part-of-speech (POS) tagging, dependency parsing, chunking, and gradient attribution. To demonstrate the effectiveness of our proposed methods, we adapt two commonly-used candidate replacement approaches to replace the selected candidate words: synonym replacement (i.e., replace with a synonym) and masked language model replacement (i.e., replace with a candidate generated from a masked language model). We conduct experiments on three corpora and systematically evaluate our proposed methods with different metrics. Experimental results and analyses show that our proposed methods can effectively corrupt NER models.

In summary, our contributions are as follows:
\begin{enumerate}[topsep=0pt,itemsep=-1ex,partopsep=1ex,parsep=1ex]
\item We investigate different methods to perturb the most informative words for generating adversarial examples to attack NER systems.
\item Experiments and analyses show that the proposed methods are more effective than strong baselines in attacking models, posing a new challenge to existing NER systems.
\end{enumerate}{}

\section{Related Work}
Adversarial attacks have been receiving increasing attention in the field of NER. Prior work in this research direction can be generally classified into two categories: i) context-aware attacks and ii) entity attacks. In the context-aware attacks, only the non-entity context words are modified. To achieve this, \citet{lin-etal-2021-rockner} proposed to perturb the original context by sampling adversarial tokens via a masked-language model. \citet{simoncini-spanakis-2021-seqattack} presented multiple modification methods to substitute, insert, swap, or delete characters and words. \citet{wang-etal-2021-textflint} studied to create adversarial samples by concatenating different sentences into a single data point. For entity attacks, the entity words are modified while the non-entity context words are kept unchanged. In particular, \citet{lin-etal-2021-rockner} exploited an external dictionary from Wikidata to find replacements for entity instances. \citet{simoncini-spanakis-2021-seqattack} studied the use of the SCPNs \citep{iyyer-etal-2018-adversarial} to generate candidate paraphrases as adversarial samples. \citet{reich-etal-2022-leveraging} proposed leveraging expert-guided heuristics to modify the entity tokens and their surrounding contexts, thereby altering their entity types as adversarial attacks. \citet{wang-etal-2021-textflint} performed adversarial attacks by swapping words or manipulating characters.

\section{Context-aware Adversarial Attack}

In this work, we propose different methods to generate adversarial samples for the purpose of auditing the model robustness of NER systems. In the following sections, we describe the two main stages involved in this process: 1) candidate selection, which aims to determine which candidate words should be replaced; and 2) candidate replacement, which aims to find the best way to replace candidate words. The pipeline of adversarial data generation is shown in Figure \ref{fig: pipeline}.

\subsection{Candidate Selection}

\begin{table*}[ht]
    \centering
    \renewcommand{\arraystretch}{1.2}
    \resizebox{0.8\linewidth}{!}{
        \begin{tabular}{ccc}
            \toprule
            \textbf{CoNLL03} & \textbf{OntoNotes5.0} & \textbf{W-NUT17} \\
            \toprule
            \includegraphics[scale=0.45]{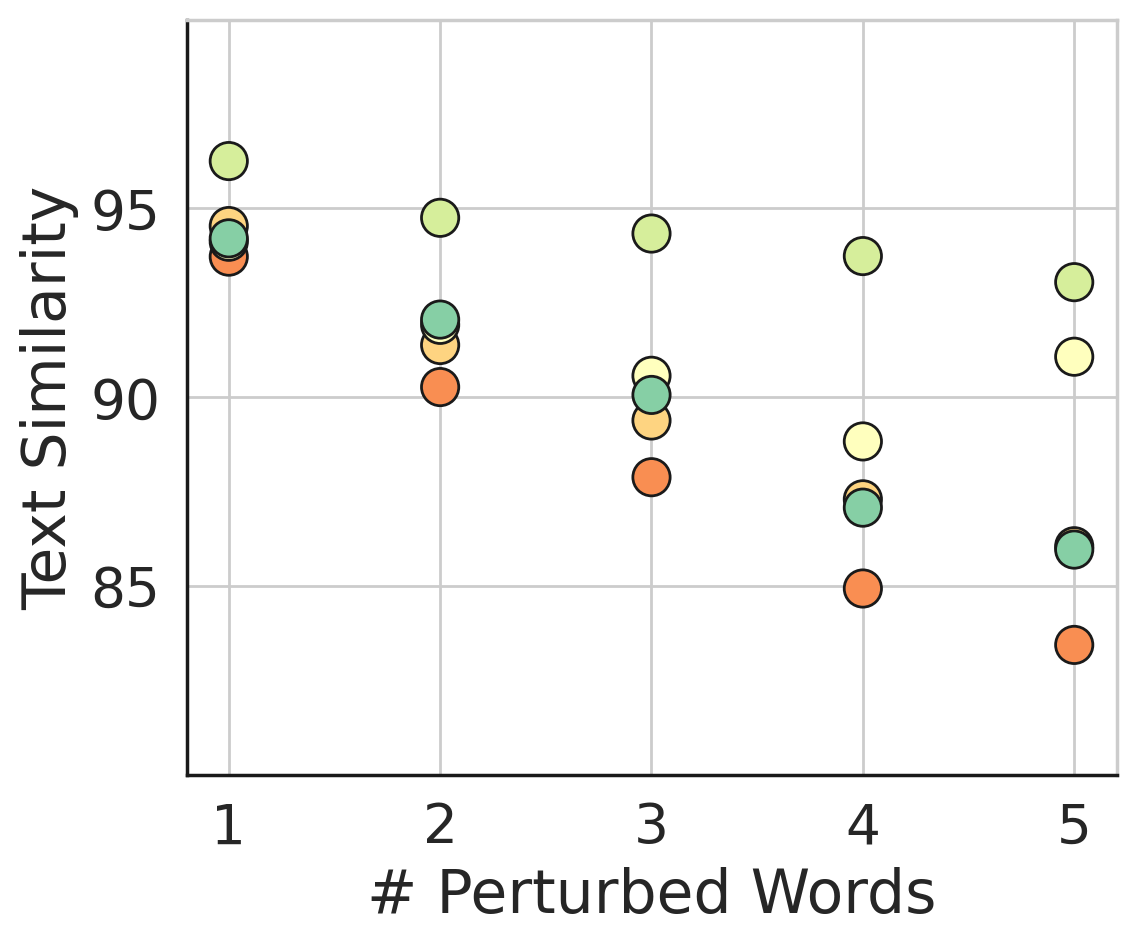} &
            \includegraphics[scale=0.45]{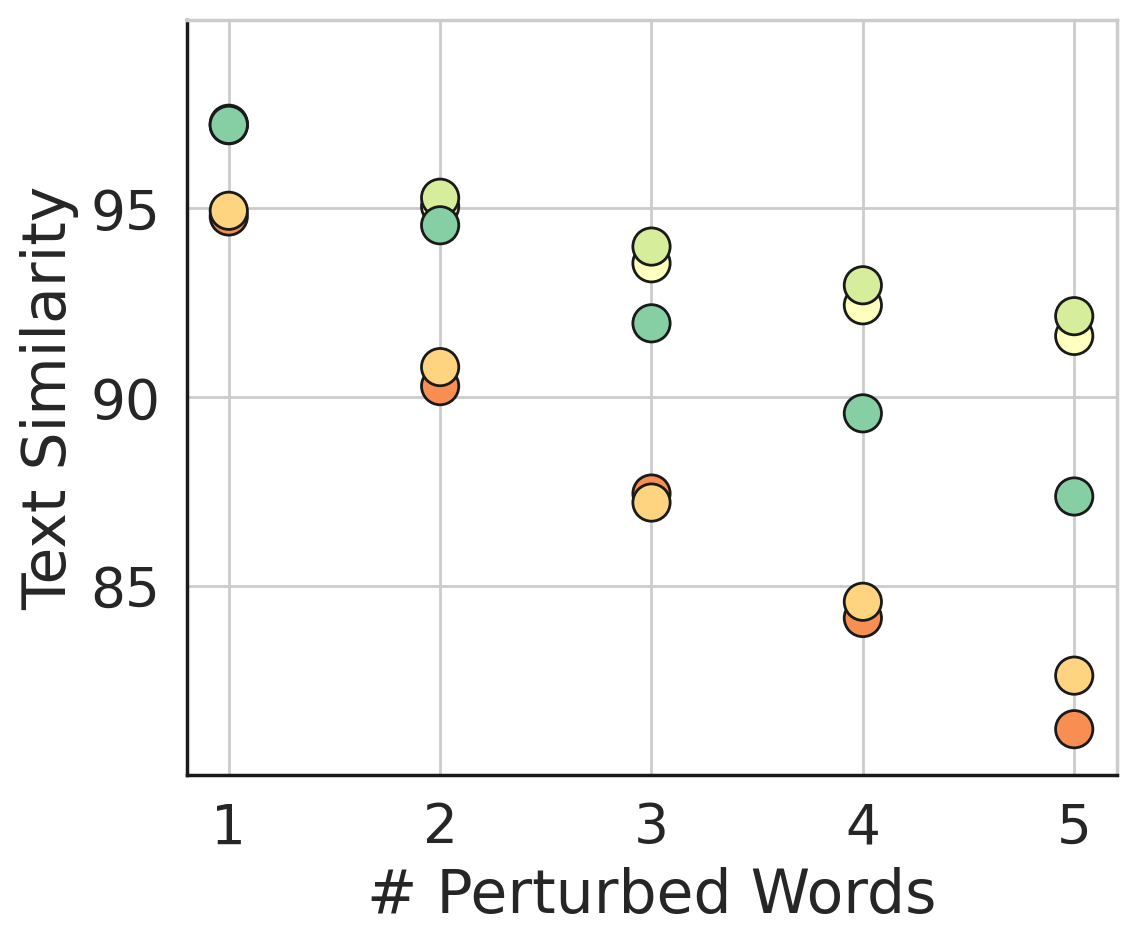} &
            \includegraphics[scale=0.45]{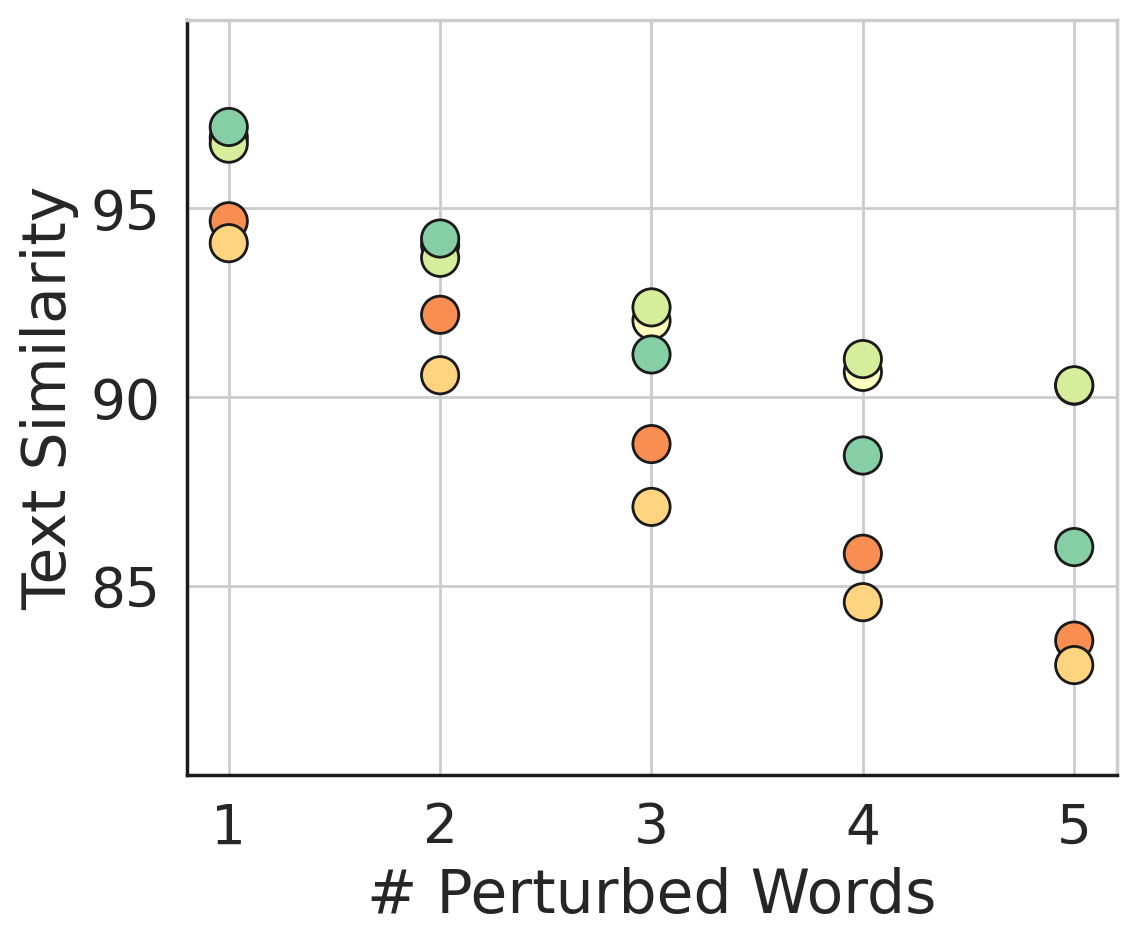} \\
            \midrule
            \includegraphics[scale=0.45]{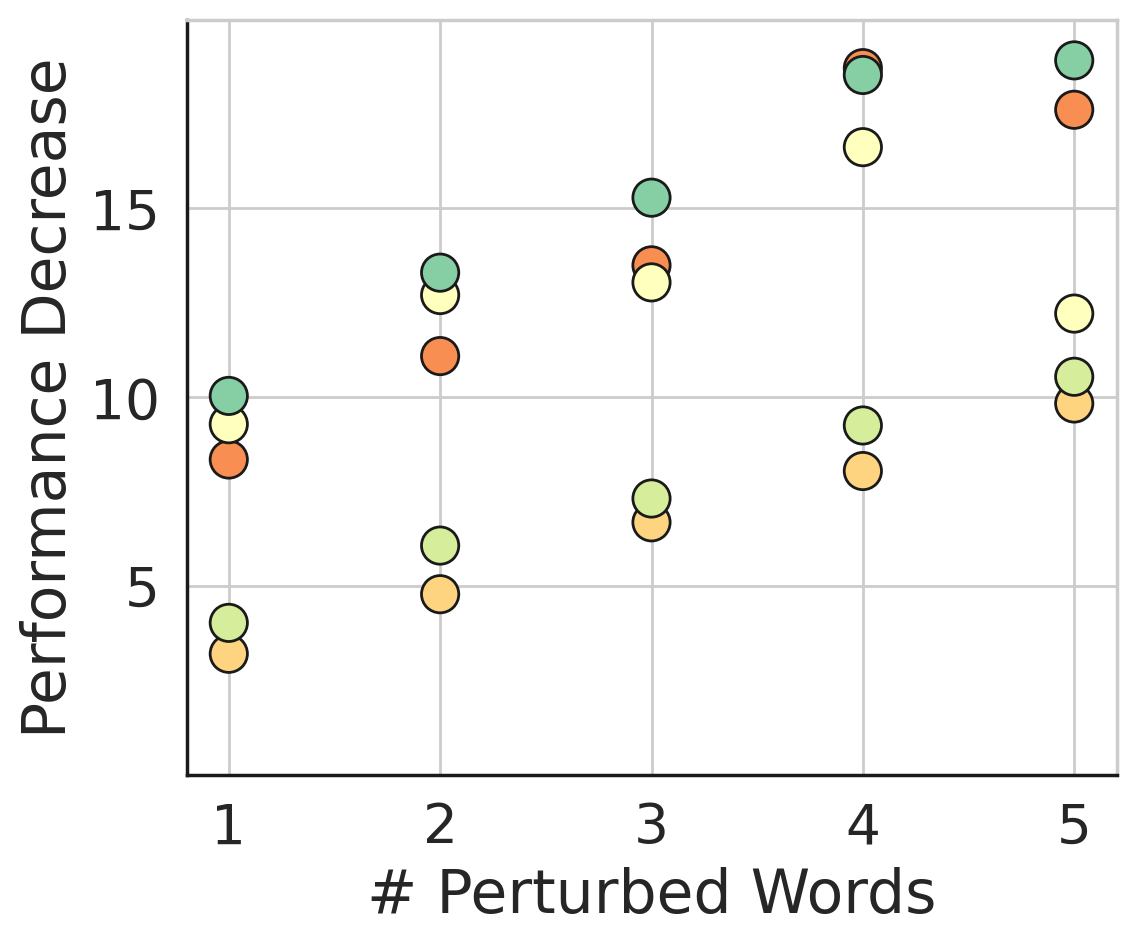} &
            \includegraphics[scale=0.45]{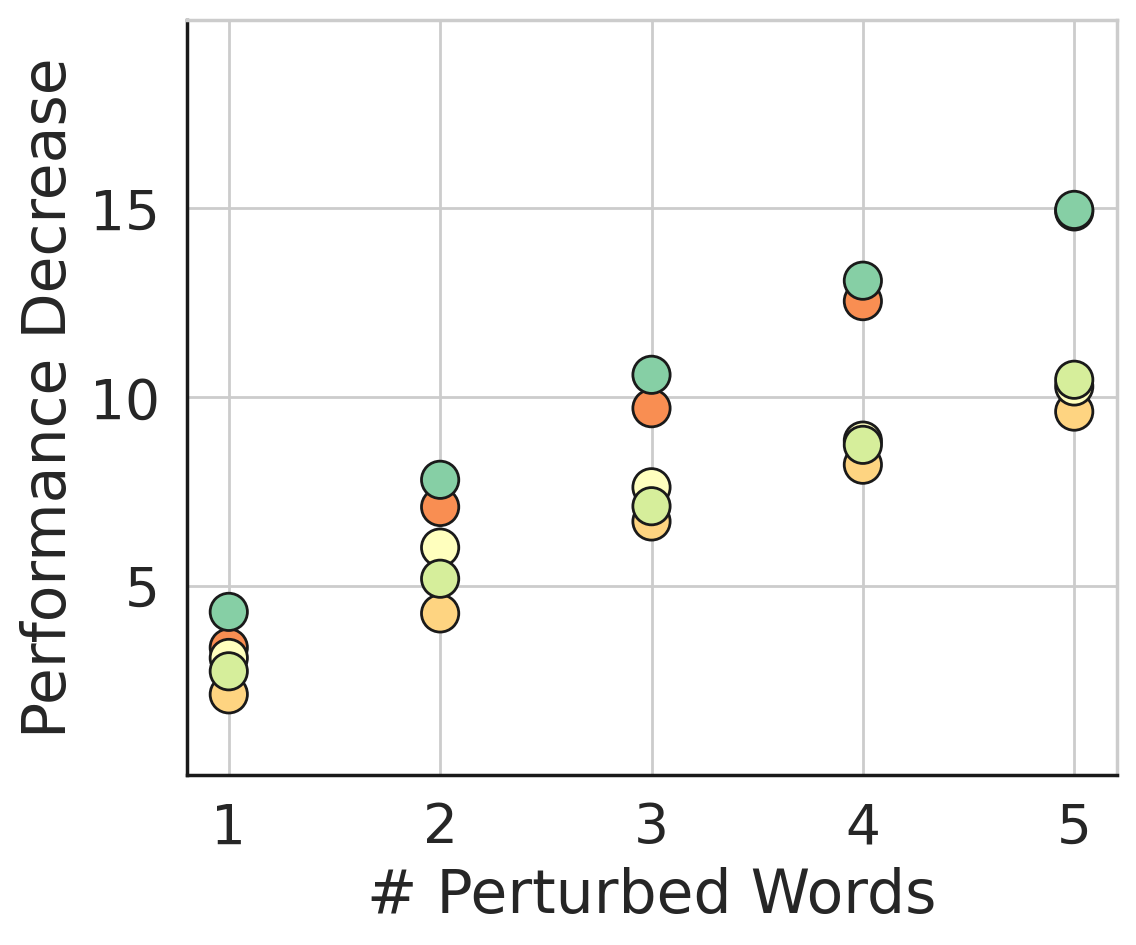} &
            \includegraphics[scale=0.45]{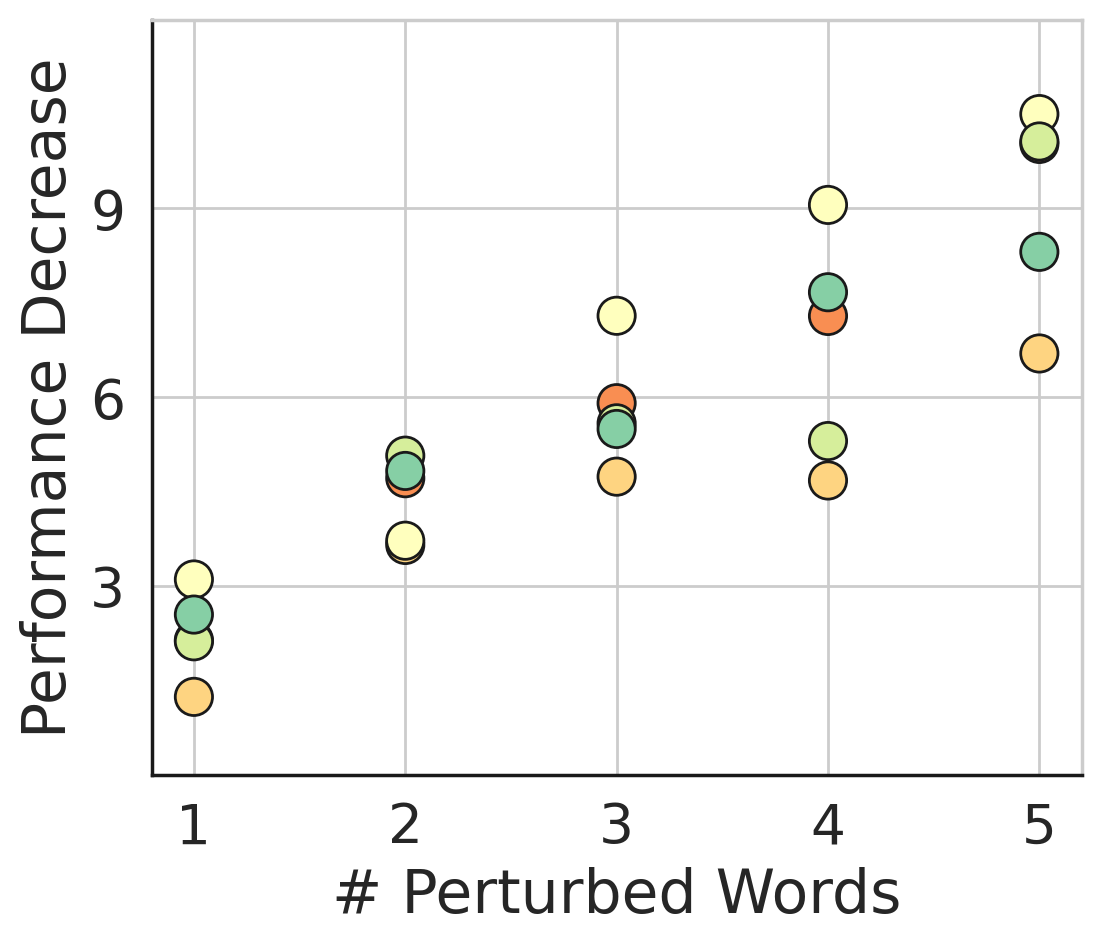} \\
            \toprule
            \multicolumn{3}{c}{\includegraphics[scale=0.5]{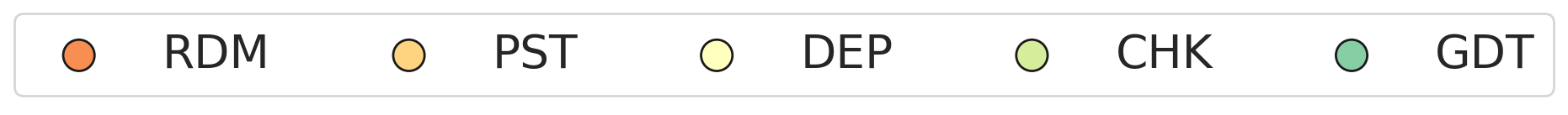}} \\
            \toprule
        \end{tabular}
    }
    \caption{Comparison between different candidate selection methods using synonym replacement. RDM, PST, DEP, CHK, GDT are short for random, POS tagging, dependency parsing, chunking, and gradient candidate selection, respectively. The x-axis denotes the number of perturbed words while the y-axis denotes the difference in F1 scores.}
    \label{tab: compare_candidate_selection}
\end{table*}

To effectively attack the model, we consider perturbing the most informative words for recognizing entities. We investigate the following automated methods to select such words as candidates:

\begin{itemize}[topsep=0pt,itemsep=-1ex,partopsep=1ex,parsep=1ex]
\item \textbf{Random (RDM)}: select non-entity words at random from the sentence as candidate words. 
\item \textbf{POS tagging (PST)}: select semantic-rich non-entity words as candidate words based on their POS tags. Here, following \citet{lin-etal-2021-rockner}, we consider selecting adjectives, nouns, adverbs, and verbs.
\item \textbf{Dependency parsing (DEP)}: select the non-entity words related to entity instances, including ascendants and descendants, as candidate words based on dependency parsing. 
\item \textbf{Chunking (CHK)}: select the non-entity words in the noun chunks that are close to entity instances as candidate words to preserve both semantic and syntactic coherence.
\item \textbf{Gradient (GDT)}: select the non-entity words according to the integral of gradients. We use Integrated Hessians \citep{DBLP:journals/jmlr/JanizekSL21} to determine the importance of non-entity words based on their feature interactions with entity instances, and select the words with higher importance scores to perturb.
\end{itemize}

To obtain linguistic features, including part-of-speech tags, dependency parsing, and chunking, for our proposed method, we use the statistical model from spaCy \footnote{\url{https://github.com/explosion/spaCy}} to process text. Then we select the candidate words to perturb based on this information. For GDT, we use the gradient of the pre-trained $\text{BERT}_\text{base}$ model \citep{devlin-etal-2019-bert} to determine the importance of each word.

\subsection{Candidate Replacement}

\begin{table*}[ht]
    \centering
    \renewcommand{\arraystretch}{1.2}
    \resizebox{0.75\linewidth}{!}{
        \begin{tabular}{cccc}
            \toprule
            & \textbf{CoNLL03} & \textbf{OntoNotes5.0} & \textbf{W-NUT17} \\
            \toprule
            \multirow{7}{*}{\textbf{\makecell[c]{Text \\ Similarity}}} &
            \multirow{7}{*}{\includegraphics[scale=0.35]{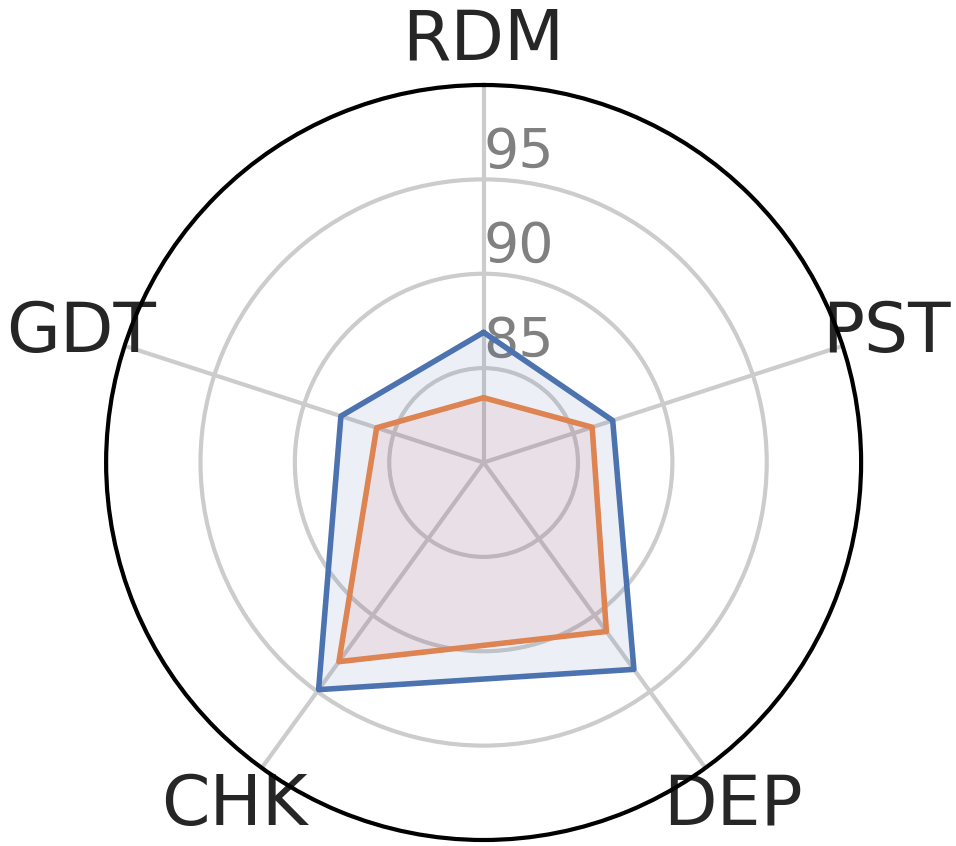}} &
            \multirow{7}{*}{\includegraphics[scale=0.35]{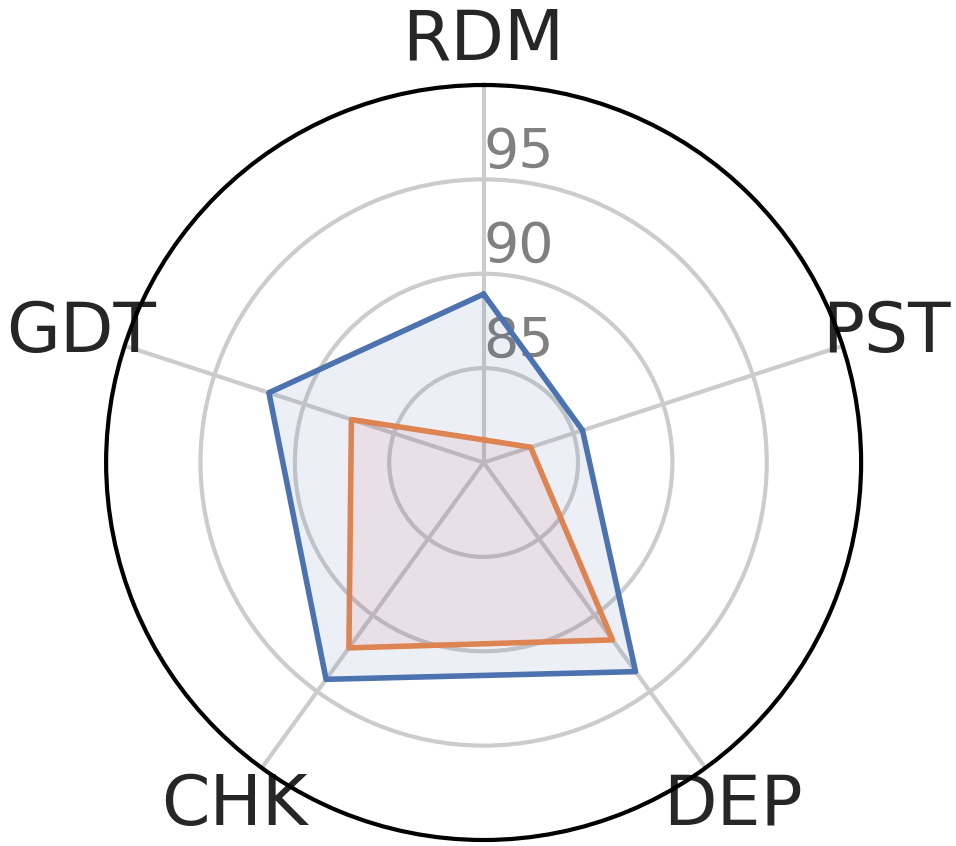}} &
            \multirow{7}{*}{\includegraphics[scale=0.35]{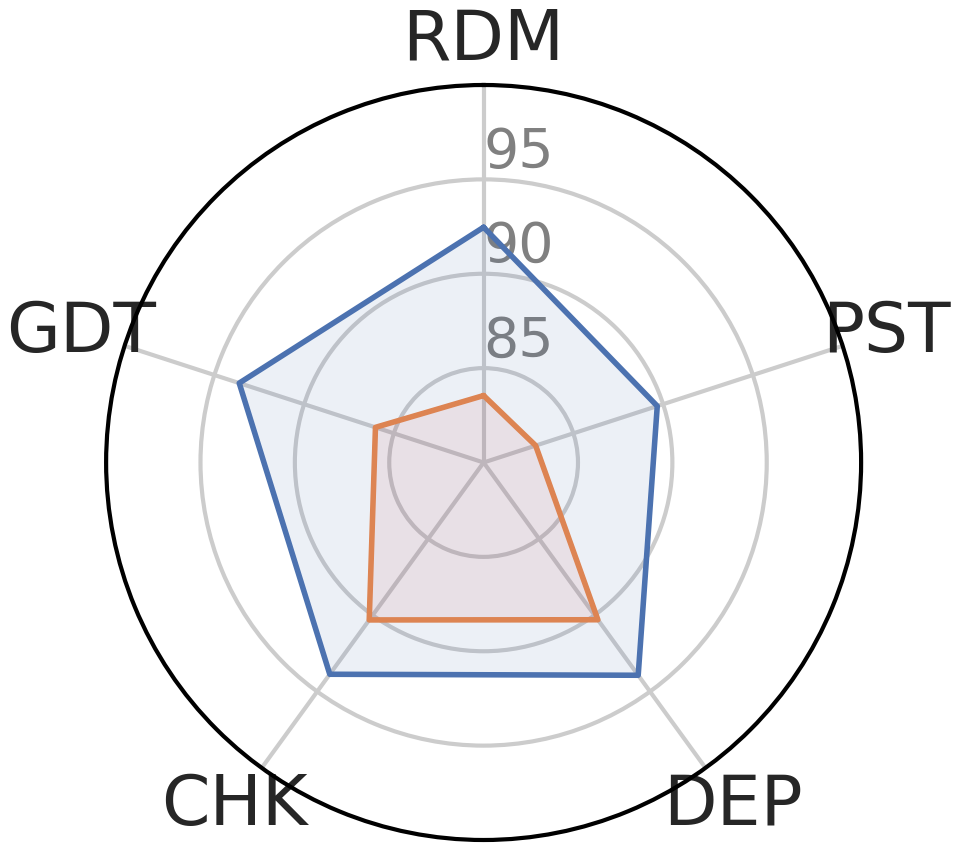}} \\
            &&&\\
            &&&\\
            &&&\\
            &&&\\
            &&&\\
            &&&\\
            \midrule
            \multirow{7}{*}{\textbf{\makecell[c]{Performance \\ Decrease}}} &
            \multirow{7}{*}{\includegraphics[scale=0.35]{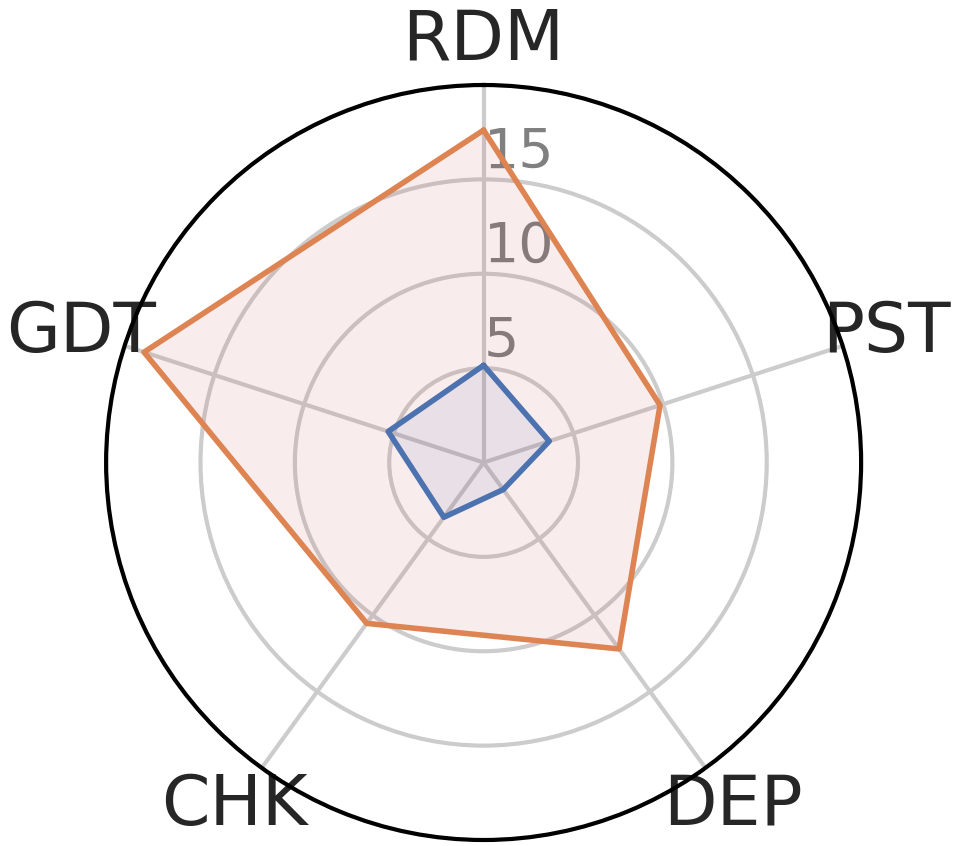}} &
            \multirow{7}{*}{\includegraphics[scale=0.35]{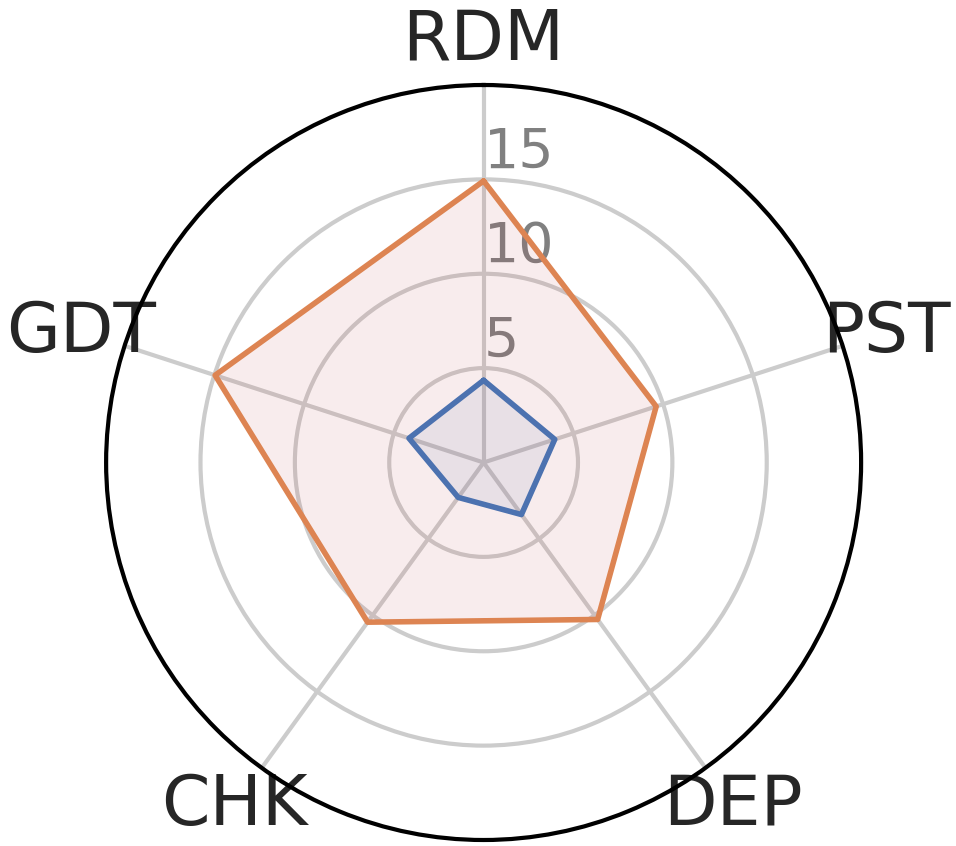}} &
            \multirow{7}{*}{\includegraphics[scale=0.35]{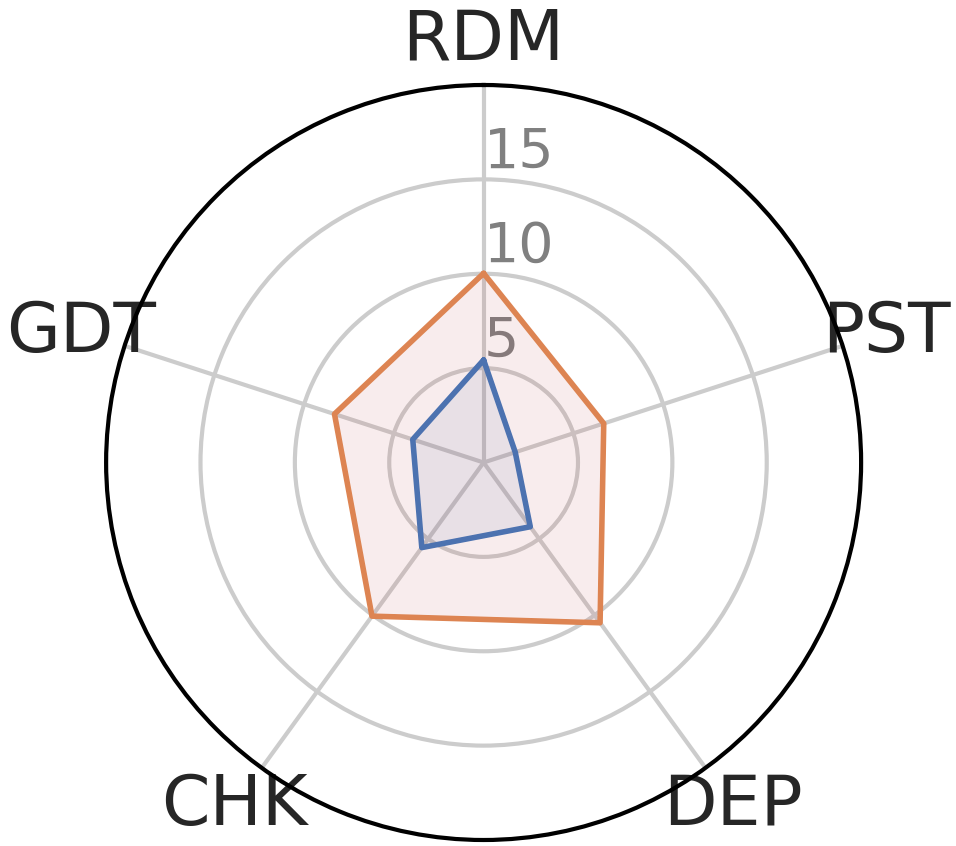}} \\
            &&&\\
            &&&\\
            &&&\\
            &&&\\
            &&&\\
            &&&\\
            \toprule
            \multicolumn{4}{c}{\includegraphics[scale=0.8]{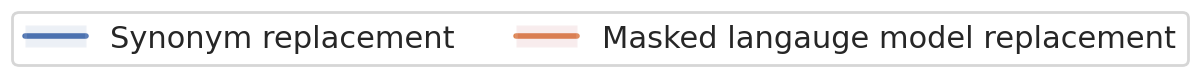}} \\
            \toprule
        \end{tabular}
    }
    \caption{Comparison between different candidate replacement methods when perturbing five words in each sentence. RDM, PST, DEP, CHK, GDT are short for random, POS tagging, dependency parsing, chunking, and gradient candidate selection, respectively.}
    \label{tab: compare_candidate_replacement}
\end{table*}

Perturbations in text at the character-level can be easily detected and defended by spell check and correction \citep{pruthi-etal-2019-combating, li-etal-2020-context-aware}. Therefore, we exclusively focus on the word-level perturbations in this work. Simply replacing a word with another one at random can lead to noisy data. For instance, in Figure \ref{fig: illustration_examples}, the label for ``Wilson" is changed from \textit{ORG} to \textit{PER} by replacing ``rackets" with ``guidance", which has a conflict with its original gold label. Therefore, to keep original labels valid, we investigate the following two approaches to replace candidate words:

\begin{itemize}[topsep=0pt,itemsep=-1ex,partopsep=1ex,parsep=1ex]
\item \textbf{Synonym Replacement}: Using synonyms to replace candidate words as adversarial samples can guarantee the preservation of text semantics and make it hard to be perceived by human investigation. We use the WordNet \citep{miller1998wordnet} dictionary to find synonyms for candidate words, and then randomly select one of them as a replacement.
\item \textbf{Masked Language Model Replacement}: The masked language model (MLM) attempts to predict the masked words in the given input sequence. In our work, we first create masks for candidate words, and then use a masked language model $\text{RoBERTa}_{\text{base}}$ \citep{DBLP:journals/corr/abs-1907-11692} to generate a replacement based on the context. This approach is capable of preserving both semantics and syntax in the generated adversarial samples.
\end{itemize}

\section{Experiments}

In this section, we present the experimental setup and results. We systematically conduct experiments to evaluate our proposed methods on three corpora with different metrics and provide analyses to better understand their effectiveness.

\subsection{Experiment Setup}

\paragraph{Datasets}
We evaluate the proposed methods on three commonly-used corpora for NER tasks, including CoNLL03 \citep{tjong-kim-sang-de-meulder-2003-introduction}, OntoNotes5.0 \citep{pradhan-etal-2013-towards}, and W-NUT17 \citep{derczynski-etal-2017-results}. The data statistics are summarized in Appendix \ref{sec: data-statistics}. 

\paragraph{Victim Model}
The victim model consists of the $\text{BERT}_\text{base}$ \citep{devlin-etal-2019-bert} as the base model and a linear layer as the classifier to assign NER tags. The details of hyper-parameters and fine-tuning are described in Appendix \ref{sec: hyperparameters-and-finetuning}.

\paragraph{Evaluation Metrics}
To examine the effectiveness of our proposed methods, we consider the following metrics for evaluation:
\begin{itemize}[topsep=0pt,itemsep=-1ex,partopsep=1ex,parsep=1ex]
\item \textbf{Textual Similarity (Sim.)}: cosine similarity between adversarial examples and the corresponding original examples using the Universal Sentence Encoder \citep{giorgi-etal-2021-declutr}. A higher textual similarity score indicates that more semantics are preserved.
\item \textbf{Performance Decrease ($\Delta$Perf.)}: the difference in F1 scores between adversarial examples and their corresponding original examples. A higher performance decrease indicates that the model makes more mistakes.
\end{itemize}{}

\subsection{Main Results} 

We compare candidate selection and replacement methods by perturbing the same number of words in the sentences. Below we present experimental results and summarize our findings:

\paragraph{Candidate Selection V.S. Metrics}
From the results in Table \ref{tab: compare_candidate_selection}, we observe that the model performance decreases rapidly under adversarial attacks. When perturbing five words in the sentence, the F1 scores decrease by 10\% \textasciitilde 20\%. Among these attack methods, GDT and RDM are more effective at deceiving the model into making wrong predictions. When performing attacks with RDM, however, the text similarity is sacrificed in exchange for a greater performance decrease, which can potentially make adversarial examples easier to detect. Additionally, it is worth noting that DEP is also effective at a slight perturbation, although it can only result in a smaller performance decrease as we increase the number of perturbed words. In terms of textual similarity and performance decrease, PST is the least effective method in most cases.

\paragraph{Candidate Replacement V.S. Metrics}
The comparison between different candidate replacement methods is shown in Table \ref{tab: compare_candidate_replacement}. In general, compared to masked language model replacement, synonym replacement can achieve a higher textual similarity, indicating that more semantics are preserved in adversarial examples. However, its performance decrease is quite limited. At a slightly lower textual similarity, masked language model replacement leads to a much larger performance decrease. Besides, both replacement methods are relatively less effective on the W-NUT17 corpus. Compared to the text from CoNLL03 and OntoNotes5.0 which is long and formal, the text from W-NUT17 is short and noisy as it contains many misspellings and grammar errors. For this reason, the model cannot rely too heavily on context when making predictions, limiting the effectiveness of adversarial attacks on this corpus.

\section{Conclusion}
In this work, we study adversarial attacks to examine the model robustness using adversarial examples. We focus on the NER task and propose context-aware adversarial attack methods to perturb the most informative words for recognizing entities. Moreover, we investigate different candidate replacement methods for generating adversarial examples. We undertake experiments on three corpora and show that the proposed methods are more effective in attacking models than strong baselines.

\section*{Limitations}
The proposed methods require linguistic knowledge (e.g., part-of-speech tags and dependency parsing) to processing the text. Most existing tools can automate this process for English. However, these tools may need to be extended to support other languages, especially for minority languages. Additionally, the proposed methods maybe not applicable with low computational resources or in real-time scenarios. 

\section*{Acknowledgements}
This work received partial support from the National Science Foundation under award number 1910192.
\bibliography{anthology,custom}

\begin{thebibliography}{22}
\expandafter\ifx\csname natexlab\endcsname\relax\def\natexlab#1{#1}\fi

\bibitem[{Belinkov and Bisk(2018)}]{DBLP:conf/iclr/BelinkovB18}
Yonatan Belinkov and Yonatan Bisk. 2018.
\newblock \href {https://openreview.net/forum?id=BJ8vJebC-} {Synthetic and
  natural noise both break neural machine translation}.
\newblock In \emph{6th International Conference on Learning Representations,
  {ICLR} 2018, Vancouver, BC, Canada, April 30 - May 3, 2018, Conference Track
  Proceedings}. OpenReview.net.

\bibitem[{Cheng et~al.(2019)Cheng, Jiang, and
  Macherey}]{cheng-etal-2019-robust}
Yong Cheng, Lu~Jiang, and Wolfgang Macherey. 2019.
\newblock \href {https://doi.org/10.18653/v1/P19-1425} {Robust neural machine
  translation with doubly adversarial inputs}.
\newblock In \emph{Proceedings of the 57th Annual Meeting of the Association
  for Computational Linguistics}, pages 4324--4333, Florence, Italy.
  Association for Computational Linguistics.

\bibitem[{Derczynski et~al.(2017)Derczynski, Nichols, van Erp, and
  Limsopatham}]{derczynski-etal-2017-results}
Leon Derczynski, Eric Nichols, Marieke van Erp, and Nut Limsopatham. 2017.
\newblock \href {https://doi.org/10.18653/v1/W17-4418} {Results of the
  {WNUT}2017 shared task on novel and emerging entity recognition}.
\newblock In \emph{Proceedings of the 3rd Workshop on Noisy User-generated
  Text}, pages 140--147, Copenhagen, Denmark. Association for Computational
  Linguistics.

\bibitem[{Devlin et~al.(2019)Devlin, Chang, Lee, and
  Toutanova}]{devlin-etal-2019-bert}
Jacob Devlin, Ming-Wei Chang, Kenton Lee, and Kristina Toutanova. 2019.
\newblock \href {https://doi.org/10.18653/v1/N19-1423} {{BERT}: Pre-training of
  deep bidirectional transformers for language understanding}.
\newblock In \emph{Proceedings of the 2019 Conference of the North {A}merican
  Chapter of the Association for Computational Linguistics: Human Language
  Technologies, Volume 1 (Long and Short Papers)}, pages 4171--4186,
  Minneapolis, Minnesota. Association for Computational Linguistics.

\bibitem[{Garg and Ramakrishnan(2020)}]{garg-ramakrishnan-2020-bae}
Siddhant Garg and Goutham Ramakrishnan. 2020.
\newblock \href {https://doi.org/10.18653/v1/2020.emnlp-main.498} {{BAE}:
  {BERT}-based adversarial examples for text classification}.
\newblock In \emph{Proceedings of the 2020 Conference on Empirical Methods in
  Natural Language Processing (EMNLP)}, pages 6174--6181, Online. Association
  for Computational Linguistics.

\bibitem[{Giorgi et~al.(2021)Giorgi, Nitski, Wang, and
  Bader}]{giorgi-etal-2021-declutr}
John Giorgi, Osvald Nitski, Bo~Wang, and Gary Bader. 2021.
\newblock \href {https://doi.org/10.18653/v1/2021.acl-long.72} {{D}e{CLUTR}:
  Deep contrastive learning for unsupervised textual representations}.
\newblock In \emph{Proceedings of the 59th Annual Meeting of the Association
  for Computational Linguistics and the 11th International Joint Conference on
  Natural Language Processing (Volume 1: Long Papers)}, pages 879--895, Online.
  Association for Computational Linguistics.

\bibitem[{Iyyer et~al.(2018)Iyyer, Wieting, Gimpel, and
  Zettlemoyer}]{iyyer-etal-2018-adversarial}
Mohit Iyyer, John Wieting, Kevin Gimpel, and Luke Zettlemoyer. 2018.
\newblock \href {https://doi.org/10.18653/v1/N18-1170} {Adversarial example
  generation with syntactically controlled paraphrase networks}.
\newblock In \emph{Proceedings of the 2018 Conference of the North {A}merican
  Chapter of the Association for Computational Linguistics: Human Language
  Technologies, Volume 1 (Long Papers)}, pages 1875--1885, New Orleans,
  Louisiana. Association for Computational Linguistics.

\bibitem[{Janizek et~al.(2021)Janizek, Sturmfels, and
  Lee}]{DBLP:journals/jmlr/JanizekSL21}
Joseph~D. Janizek, Pascal Sturmfels, and Su{-}In Lee. 2021.
\newblock \href {http://jmlr.org/papers/v22/20-1223.html} {Explaining
  explanations: Axiomatic feature interactions for deep networks}.
\newblock \emph{J. Mach. Learn. Res.}, 22:104:1--104:54.

\bibitem[{Jia and Liang(2017)}]{jia-liang-2017-adversarial}
Robin Jia and Percy Liang. 2017.
\newblock \href {https://doi.org/10.18653/v1/D17-1215} {Adversarial examples
  for evaluating reading comprehension systems}.
\newblock In \emph{Proceedings of the 2017 Conference on Empirical Methods in
  Natural Language Processing}, pages 2021--2031, Copenhagen, Denmark.
  Association for Computational Linguistics.

\bibitem[{Kingma and Ba(2015)}]{DBLP:journals/corr/KingmaB14}
Diederik~P. Kingma and Jimmy Ba. 2015.
\newblock \href {http://arxiv.org/abs/1412.6980} {Adam: {A} method for
  stochastic optimization}.
\newblock In \emph{3rd International Conference on Learning Representations,
  {ICLR} 2015, San Diego, CA, USA, May 7-9, 2015, Conference Track
  Proceedings}.

\bibitem[{Li et~al.(2020)Li, Liu, and Huang}]{li-etal-2020-context-aware}
Xiangci Li, Hairong Liu, and Liang Huang. 2020.
\newblock \href {https://doi.org/10.18653/v1/2020.findings-emnlp.37}
  {Context-aware stand-alone neural spelling correction}.
\newblock In \emph{Findings of the Association for Computational Linguistics:
  EMNLP 2020}, pages 407--414, Online. Association for Computational
  Linguistics.

\bibitem[{Liang et~al.(2018)Liang, Li, Su, Bian, Li, and
  Shi}]{DBLP:conf/ijcai/0002LSBLS18}
Bin Liang, Hongcheng Li, Miaoqiang Su, Pan Bian, Xirong Li, and Wenchang Shi.
  2018.
\newblock \href {https://doi.org/10.24963/ijcai.2018/585} {Deep text
  classification can be fooled}.
\newblock In \emph{Proceedings of the Twenty-Seventh International Joint
  Conference on Artificial Intelligence, {IJCAI} 2018, July 13-19, 2018,
  Stockholm, Sweden}, pages 4208--4215. ijcai.org.

\bibitem[{Lin et~al.(2021)Lin, Gao, Yan, Moreno, and
  Ren}]{lin-etal-2021-rockner}
Bill~Yuchen Lin, Wenyang Gao, Jun Yan, Ryan Moreno, and Xiang Ren. 2021.
\newblock \href {https://doi.org/10.18653/v1/2021.emnlp-main.302} {{R}ock{NER}:
  A simple method to create adversarial examples for evaluating the robustness
  of named entity recognition models}.
\newblock In \emph{Proceedings of the 2021 Conference on Empirical Methods in
  Natural Language Processing}, pages 3728--3737, Online and Punta Cana,
  Dominican Republic. Association for Computational Linguistics.

\bibitem[{Liu et~al.(2019)Liu, Ott, Goyal, Du, Joshi, Chen, Levy, Lewis,
  Zettlemoyer, and Stoyanov}]{DBLP:journals/corr/abs-1907-11692}
Yinhan Liu, Myle Ott, Naman Goyal, Jingfei Du, Mandar Joshi, Danqi Chen, Omer
  Levy, Mike Lewis, Luke Zettlemoyer, and Veselin Stoyanov. 2019.
\newblock \href {http://arxiv.org/abs/1907.11692} {Roberta: {A} robustly
  optimized {BERT} pretraining approach}.
\newblock \emph{CoRR}, abs/1907.11692.

\bibitem[{Miller(1998)}]{miller1998wordnet}
George~A Miller. 1998.
\newblock \emph{WordNet: An electronic lexical database}.
\newblock MIT press.

\bibitem[{Pradhan et~al.(2013)Pradhan, Moschitti, Xue, Ng, Bj{\"o}rkelund,
  Uryupina, Zhang, and Zhong}]{pradhan-etal-2013-towards}
Sameer Pradhan, Alessandro Moschitti, Nianwen Xue, Hwee~Tou Ng, Anders
  Bj{\"o}rkelund, Olga Uryupina, Yuchen Zhang, and Zhi Zhong. 2013.
\newblock \href {https://aclanthology.org/W13-3516} {Towards robust linguistic
  analysis using {O}nto{N}otes}.
\newblock In \emph{Proceedings of the Seventeenth Conference on Computational
  Natural Language Learning}, pages 143--152, Sofia, Bulgaria. Association for
  Computational Linguistics.

\bibitem[{Pruthi et~al.(2019)Pruthi, Dhingra, and
  Lipton}]{pruthi-etal-2019-combating}
Danish Pruthi, Bhuwan Dhingra, and Zachary~C. Lipton. 2019.
\newblock \href {https://doi.org/10.18653/v1/P19-1561} {Combating adversarial
  misspellings with robust word recognition}.
\newblock In \emph{Proceedings of the 57th Annual Meeting of the Association
  for Computational Linguistics}, pages 5582--5591, Florence, Italy.
  Association for Computational Linguistics.

\bibitem[{Reich et~al.(2022)Reich, Chen, Agrawal, Zhang, and
  Yang}]{reich-etal-2022-leveraging}
Aaron Reich, Jiaao Chen, Aastha Agrawal, Yanzhe Zhang, and Diyi Yang. 2022.
\newblock \href {https://doi.org/10.18653/v1/2022.findings-acl.154} {Leveraging
  expert guided adversarial augmentation for improving generalization in named
  entity recognition}.
\newblock In \emph{Findings of the Association for Computational Linguistics:
  ACL 2022}, pages 1947--1955, Dublin, Ireland. Association for Computational
  Linguistics.

\bibitem[{Simoncini and Spanakis(2021)}]{simoncini-spanakis-2021-seqattack}
Walter Simoncini and Gerasimos Spanakis. 2021.
\newblock \href {https://doi.org/10.18653/v1/2021.emnlp-demo.35}
  {{S}eq{A}ttack: {O}n adversarial attacks for named entity recognition}.
\newblock In \emph{Proceedings of the 2021 Conference on Empirical Methods in
  Natural Language Processing: System Demonstrations}, pages 308--318, Online
  and Punta Cana, Dominican Republic. Association for Computational
  Linguistics.

\bibitem[{Tjong Kim~Sang and
  De~Meulder(2003)}]{tjong-kim-sang-de-meulder-2003-introduction}
Erik~F. Tjong Kim~Sang and Fien De~Meulder. 2003.
\newblock \href {https://aclanthology.org/W03-0419} {Introduction to the
  {C}o{NLL}-2003 shared task: Language-independent named entity recognition}.
\newblock In \emph{Proceedings of the Seventh Conference on Natural Language
  Learning at {HLT}-{NAACL} 2003}, pages 142--147.

\bibitem[{Wallace et~al.(2019)Wallace, Feng, Kandpal, Gardner, and
  Singh}]{wallace-etal-2019-universal}
Eric Wallace, Shi Feng, Nikhil Kandpal, Matt Gardner, and Sameer Singh. 2019.
\newblock \href {https://doi.org/10.18653/v1/D19-1221} {Universal adversarial
  triggers for attacking and analyzing {NLP}}.
\newblock In \emph{Proceedings of the 2019 Conference on Empirical Methods in
  Natural Language Processing and the 9th International Joint Conference on
  Natural Language Processing (EMNLP-IJCNLP)}, pages 2153--2162, Hong Kong,
  China. Association for Computational Linguistics.

\bibitem[{Wang et~al.(2021)Wang, Liu, Gui, Zhang, Zou, Zhou, Ye, Zhang, Zheng,
  Pang, Wu, Li, Zhang, Ma, Fei, Cai, Zhao, Hu, Yan, Tan, Hu, Bian, Liu, Qin,
  Zhu, Xing, Fu, Zhang, Peng, Zheng, Zhou, Wei, Qiu, and
  Huang}]{wang-etal-2021-textflint}
Xiao Wang, Qin Liu, Tao Gui, Qi~Zhang, Yicheng Zou, Xin Zhou, Jiacheng Ye,
  Yongxin Zhang, Rui Zheng, Zexiong Pang, Qinzhuo Wu, Zhengyan Li, Chong Zhang,
  Ruotian Ma, Zichu Fei, Ruijian Cai, Jun Zhao, Xingwu Hu, Zhiheng Yan, Yiding
  Tan, Yuan Hu, Qiyuan Bian, Zhihua Liu, Shan Qin, Bolin Zhu, Xiaoyu Xing,
  Jinlan Fu, Yue Zhang, Minlong Peng, Xiaoqing Zheng, Yaqian Zhou, Zhongyu Wei,
  Xipeng Qiu, and Xuanjing Huang. 2021.
\newblock \href {https://doi.org/10.18653/v1/2021.acl-demo.41} {{T}ext{F}lint:
  Unified multilingual robustness evaluation toolkit for natural language
  processing}.
\newblock In \emph{Proceedings of the 59th Annual Meeting of the Association
  for Computational Linguistics and the 11th International Joint Conference on
  Natural Language Processing: System Demonstrations}, pages 347--355, Online.
  Association for Computational Linguistics.

\end{thebibliography}
\bibliographystyle{acl_natbib}

\appendix

\section{Data Statistics}
\label{sec: data-statistics}
Table \ref{tab: data_statistics} shows data statistics of the NER datasets we used in our experiments:

\begin{table}[ht]
    \centering
    \renewcommand{\arraystretch}{1.2}
    \resizebox{\linewidth}{!}{
        \begin{tabular}{l|ccc}
            \hline
            \bf Split & \bf CoNLL03 & \bf OntoNotes5.0 & \bf W-NUT17 \\
            \hline
            Train       & 14,041    & 115,812   & 3,394 \\
            Validation  & 3,250     & 15,680    & 1,009 \\
            Test        & 3,453     & 12,217    & 1,287 \\
            \hline
            Total       & 20,744    & 143,709   & 5,690  \\
            \hline
        \end{tabular}
    }
    \caption{Data Statistics of CoNLL03, OntoNotes5.0 and W-NUT17 corpus.}
    \label{tab: data_statistics}
\end{table}

\section{Hyper-parameters and Fine-tuning}
\label{sec: hyperparameters-and-finetuning}
For the victim model, we use the $\text{BERT}_\text{base}$ \citep{devlin-etal-2019-bert} without changing any hyper-parameters. The learning rate is set to 5e-5 and the training batch size is set to 8. We train the model using the Adam optimizer \citep{DBLP:journals/corr/KingmaB14} with a weight decay 0.01 for 10 epochs on CoNLL03 and OntoNotes5.0 data and 20 epochs on W-NUT17 data. For the hardware, we use 8 NVIDIA V100 GPUs with a memory of 24GB.

\end{document}